\documentclass[runningheads,a4paper]{llncs}

\usepackage{algorithm}
\usepackage{algorithmic} 
\usepackage{bm}
\usepackage{amssymb}
\usepackage{graphicx}

\usepackage{url}
\urldef{\mailsa}\path|qiuyangliu2014@gmail.com |
\urldef{\mailsb}\path|slsun@cs.ecnu.edu.cn|

\begin{document}

\mainmatter  

\title{Multi-view Regularized Gaussian Processes}

\titlerunning{Multi-view Regularized Gaussian Processes}

%
%
\author{Qiuyang Liu%
\and Shiliang Sun}
%

\institute{Department of Computer Science and Technology, East China Normal University,\\
3663 North Zhongshan Road, Shanghai 200062, P. R. China\\
\mailsa\\
\mailsb}


%
%

\maketitle

\begin{abstract}
Gaussian processes (GPs) have been proven to be powerful tools in various areas of
machine learning. However, there are very few applications of GPs in the scenario of
multi-view learning. In this paper, we present a new GP model for multi-view learning. Unlike
existing methods, it combines multiple views by regularizing marginal likelihood
with the consistency among the posterior distributions of latent functions from
different views. Moreover, we give a general point selection scheme for multi-view learning and
improve the proposed model by this criterion.
Experimental results on multiple real world data sets have verified the
effectiveness of the proposed model and witnessed the performance improvement through
employing this novel point selection scheme.
\end{abstract}

\section{Introduction}

Gaussian processes (GPs) \cite{ras} are flexible and popular Bayesian nonparametric
tools for probabilistic modeling.
Without giving concrete functional forms, they can be employed to define
distributions over functions. As a kind of effective probabilistic models, they provide
estimations of the uncertainty of predictions.
With many convenient properties, GPs are widely used for various applications in
machine learning and statistics. For instance, GPs have progressed dramatically in
semi-supervised learning \cite{law,zhu,sin}, active learning \cite{kra,zhou}, multi-task learning \cite{yu,bon}, reinforcement learning \cite{eng,rasa},
and time series modeling \cite{dam,zha}.

Standard GPs only deal with single view data. However, in real world, many data
involve multi-view information, which may come from different feature extractors or
different domains. For example, in web-page classification, a web-page can be
described by its content and its hyperlink structure. In image classification, an
image can be represented by its color, texture, shape, and so on.
Therefore, recently, multi-view learning has aroused wide concern in machine
learning.
There are increasing number of algorithms proposed for multi-view learning, which can
mainly be divided into two major categories \cite{sun}: co-training style algorithm
\cite{blu,sun2} and
co-regularization style algorithms \cite{yus,jas}.
However, GPs, as efficient and elegant methods in machine learning, have very few
applications in multi-view learning \cite{yus,sho}.
Our work extends the GPs to the scenario of multi-view learning.

Existing multi-view learning methods which involve the GPs can be classified into two
groups: Bayesian co-training \cite{yus} and subspace learning \cite{sho,xuc}. The
Bayesian co-training approach \cite{yus} is a Bayesian undirected graphical model,
which pays attention to semi-supervised multi-view learning. The conditional independence between the output $y$ of each
data and latent functions $f_j$ for each view is ensured by involving a latent function $f_c$ \cite{xu}. On the other hand, the subspace
learning methods \cite{sho,xuc} use the GPs as tools to construct a latent variable model
which could tackle the task of non-linear dimensional reduction.
In contrast to those existing methods, inspired by the thought of co-regularization,
our work focuses on directly extending the GPs to the context of the multi-view
learning via the posterior consistency regularization, leading to elegant inference and
optimization.

Our method models the classifier of each separated view as a Gaussian process. We optimize
hyperparameters of the GPs by maximizing weighted average of marginal likelihood on
each view and minimizing the discrepancy among the posterior distributions of the latent
function on each view.
As the Kullback-Leibler (KL) divergence \cite{joy} is a frequently used measure for describing
the difference between two probability distributions, we employ it to characterize
the discrepancy among the posterior distributions.
Moreover, as data sets in real word are complex and may be seriously contaminated by
noises, the sufficiency assumption, i.e., each view is sufficient for classification
on its own, and the compatibility assumption, i.e., the target functions of all the
views predict the same labels with a high probability \cite{xu}, may fail in some cases. In
consideration of these situations, we improve our model by a selective
regularization idea, which is different from previous multi-view methods. In the
experiments, we have compared the improved method with the original model to verify the effectiveness of the
idea of the selective regularization on real word data sets.

The highlights of our work can be summarized as follows.
First, we present a new GP model for multi-view learning, which extends
the GPs to the scenario of multi-view learning by simple and elegant posterior
consistency regularization.
Secondly, our models automatically learn which views of the data should be trusted more when
predicting class labels.
Finally, we give a general point selection scheme for multi-view learning to deal
with the situations where the sufficiency and compatibility assumptions fail, and
propose the multi-view GPs with selective posterior consistency inspired by this
criterion.

The remainder of this paper is organized as follows.
Section 2 reviews the Gaussian processes.
In Section 3, we present the multi-view GPs with posterior consistency (MvGP1),
our first algorithm, covering the principles and detailed inference and
learning in the proposed model.
Moreover, we improve the MvGP1 to the multi-view GPs with selective posterior
consistency (MvGP2) based on a general idea of the consistent set in Section 4.
Experimental results are provided in Section 5.
Finally, we conclude this paper and discuss the future work in Section 6.

\section{Gaussian Processes}
This section briefly reviews the Gaussian process (GP) model.

GPs are powerful tools for classification and regression.
A Gaussian process is a collection of random variables, any finite number of which have a joint Gaussian distribution \cite{ras}.
The GP is widely used to describe a distribution over functions, and can be completely specified by its mean function and covariance function.
Formally, suppose that the training set has $N$ examples $\{(\bm x_i,y_i)\}_{i=1}^N$,
where $\bm x_i\in R^M$ is the $i$th input, and $y_i\in R$ is the corresponding label.
Denote $ \bm X=[\bm x_1,\bm x_2,...,\bm x_N]^\mathbf{T}$, and $\bm y=[y_1,y_2,...,y_N]^\mathbf{T}$.
Following standard notations for GPs,
the prior distribution for the latent functions $\bm f$ is assumed to be Gaussian,
 $\bm f|\bm X   \sim \mathcal{N}(\bm 0,\bm K)$,
with a zero mean and a covariance matrix $\bm K$, whose element $K_{ij}$ is determined by the covariance function $k(\bm x_i,\bm x_j)$.
Diverse covariance functions can be employed in GPs. In this paper, we select a commonly used covariance function,
the squared exponential kernel,
\begin{equation}
\label{eq1}
\ k(\bm x_i, \bm x_j) = s_f^2\exp(-\frac{1}{2l^2}\sum_{d=1}^N(x_{id}-x_{jd})^2),
\end{equation}
where $s_f^2$ is the signal variance, and $l$ is the length-scale of the covariance.

The Gaussian likelihood for regression can be written as
$\bm y|\bm f  \sim \mathcal{N}(\bm f, \sigma^2 \bm I)$ ,
and after integrating out the hidden variables, the marginal likelihood is
$\bm y| \bm X   \sim \mathcal{N}(0, \bm K+ \sigma^2 \bm I )$.

Under these settings, the posterior of the latent functions should be
\begin{equation}
\ \bm f|\bm y   \sim \mathcal{N}(\bm \mu,\bm \Sigma) ,
\end{equation}
where $\bm \mu  =\bm K(\bm K+\sigma^2\bm I)^{-1}\bm y$ and
$\bm \Sigma =\bm K-\bm K(\bm K+\sigma^2\bm I)^{-1}\bm K$
are the mean and the covariance of the posterior distribution, respectively.

We use $\Theta$ to denote the hyperparameters in the Gaussian process regression model, that is, $\Theta=\{s_f^2, l, \sigma^2 \}$. These hyperparameters can be obtained by generalized maximum likelihood.
In generalized maximum likelihood, we calculate the negative logarithmic marginal likelihood of the samples,
$ L(\Theta)= -\log p(\bm y| \bm X,\Theta)$,
and then minimize
$ L(\Theta)$  with respect to $\Theta$.

For a new point $\bm x^*$, the prediction is also Gaussian,
\begin{equation}
f_*|\bm X, \bm y, \bm x^* \sim \mathcal{N}(\bar{f_*}, \rm cov (f_*)),
\end{equation}
where
$\bar{f_*} = {\bm k_*}^ \mathbf{ T }[\bm K+\sigma^2\bm I]^{-1}\bm y$,
${\rm cov} (f_*) =k(\bm x_*,\bm x_*)-{\bm k_*}^ \mathbf{ T }[\bm K+\sigma^2\bm I]^{-1}\bm k_*$.
Here, $k$ is the covariance function, and $\bm{k_*}$ is the vector of covariance function values between $\bm x_*$ and the training data $\bm X$.

Standard GPs only handle single view data. However, collected data sets in the real world can often be represented by multiple views which may come from different feature extractors or various measurement modalities. As GPs are popular tools in machine learning, we propose to develop the GPs to multi-view learning.

\section{Multi-view GPs with Posterior Consistency}
In this section, we present the formulation of the multi-view GPs with posterior consistency (MvGP1) and show the corresponding inference and optimization in the proposed model. We first pay attention to two views learning tasks, and then give illustrations about the extensions to the scenario having more than two views.

\subsection{Model Representation}
Assume that the two views training set $\bm D$ has $N$ examples $\{(\bm x_i, \bm z_i, y_i)\}_{i=1}^N $, where $\bm x_i\in R^{M_1}$ is the $i$th input on the first view, $\bm z_i\in R^{M_2}$ is the $i$th input on the second view,  and $y_i\in \{+1,-1\}$ is the corresponding label.
Denote $\bm X=[\bm x_1,...,\bm x_N]^\mathbf{T}$, $\bm Z=[\bm z_1,...,\bm z_N]^\mathbf{T}$, and  $\bm{y}=[y_1,...,y_N]^\mathbf{T}$.

First, on account of leveraging the information in the separated single view, we simply assume that each view of data is modeled by a GP.
That is, the prior distribution for the latent functions $\bm f_1$ on the first view and $\bm f_2$ on the second view are supposed to be Gaussian,
i.e. $ p(\bm f_1|\bm X)= \mathcal{N}(\bm 0,\bm K_1)$, and $ p(\bm f_2|\bm Z)= \mathcal{N}(\bm 0,\bm K_2)$,
where $\bm K_1$ and $\bm K_2$ are covariance matrixes determined by the corresponding covariance functions of two views, respectively.
In our model, the covariance function is the squared exponential kernel as mentioned in (\ref{eq1}).
Although the Gaussian noise model is originally developed for regression, it has also been proved effective for classification, and its
performance is typically comparable to the more complex probit and logit likelihood models used in classification problems \cite{kap}.
Therefore, we use the Gaussian regression likelihood for our classification task to enjoy the elegant exact inference.
The Gaussian likelihood for regression on the first view is $ p(\bm y|\bm f_1)= \mathcal{N}(\bm f_1, \sigma_1^2\bm I)$,
and the likelihood on the second view is $ p(\bm y|\bm f_2)= \mathcal{N}(\bm f_2, \sigma_2^2 \bm I)$. Secondly, we also need leverage the consistence between two views. The KL divergence \cite{joy} can measure the discrepancy between two distributions. Thus, we use the KL divergence between the posterior distributions on two views to regularize the objective function in MvGP1, enforcing the consistence between two views.

Suppose the posterior distribution of the latent function $\bm f_1$ on the first view is
\begin{equation}
\ p_1 = p(\bm f_1|\bm X, \bm y) =\mathcal{N}(\bm \mu_1,\bm \Sigma_1),
\end{equation}
and
the posterior distribution of the latent function $\bm f_2$ on the second view is
\begin{equation}
\ p_2 = p(\bm f_2|\bm Z, \bm y)=\mathcal{N}(\bm \mu_2,\bm \Sigma_2).
\end{equation}

Based on the above setting, our objective function of MvGP1 is
\begin{equation}
\label{eq4}
\min \L_1=\min \{-[a\log p(\bm y|\bm X)+(1-a)\log p(\bm y|\bm Z)]+\frac{b}{2}[KL(p_1||p_2)+KL(p_2||p_1)]\},
\end{equation}
where
 \begin{eqnarray}
\log p(\bm y|\bm X) & = - \frac{1}{2}\bm y^ \mathbf{ T }(\bm K_1+\sigma_1^2 \bm I)^{-1} \bm y-\frac{1}{2}\log|\bm K_1+\sigma_1^2 \bm I|-\frac{N}{2}\log2\pi
\end{eqnarray}
is the marginal likelihood on the first view,
\begin{eqnarray}
\log p(\bm y|\bm Z) & =  - \frac{1}{2}\bm y^ \mathbf{ T }(\bm K_2+\sigma_2^2\bm I)^{-1}\bm y -\frac{1}{2}\log|\bm K_2+\sigma_2^2 \bm I|-\frac{N}{2}\log2\pi
\end{eqnarray}
is the marginal likelihood on the second view,
and the KL divergences between the posterior distributions $\bm f_1$ and $\bm f_2$ are
\begin{eqnarray}
\label{eq2}
\ KL(p_1||p_2)& =  \frac{1}{2} [\log|\bm \Sigma_2|-\log|\bm \Sigma_1|+tr(\bm \Sigma_2^{-1}\bm \Sigma_1)\nonumber \\
&+(\bm \mu_2-\bm \mu_1)^ \mathbf{ T }\bm \Sigma_2^{-1}(\bm \mu_2-\bm \mu_1)-N],
\end{eqnarray}
and
\begin{eqnarray}
\label{eq3}
\ KL(p_2||p_1)& =  \frac{1}{2} [\log|\bm \Sigma_1|-\log|\bm \Sigma_2|+tr(\bm \Sigma_1^{-1}\bm \Sigma_2) \nonumber \\
& +(\bm \mu_1-\bm \mu_2)^ \mathbf{ T }\bm \Sigma_1^{-1}(\bm \mu_1-\bm \mu_2)-N] .
\end{eqnarray}
Since the KL divergence is not a symmetrical quantity, that is, $KL(p_1||p_2)\neq KL(p_2||p_1)$, and we have no general method to determine which one is better for measuring the discrepancy between the two posterior distribution $p_1$ and $p_2$,
we construct a symmetrical quantity based on the above KL divergences, i.e., $\frac{1}{2}[KL(p_1||p_2)+ KL(p_2||p_1)]$.

Parameters $\bm \mu_1$, $\bm \mu_2$, $\bm \Sigma_1$, and $\bm \Sigma_2$ in (\ref{eq2}) and (\ref{eq3}) are provided as follows.
The mean and covariance of the posterior distribution for the latent function $\bm f_1$ are
$\bm \mu_1  =\bm K_1(\bm K_1+\sigma_1^2\bm I)^{-1}\bm y$ and $\bm \Sigma_1 = \bm K_1- \bm K_1(\bm K_1+\sigma_1^2\bm I)^{-1}\bm K_1$.
The mean and covariance of the posterior distribution for the latent function $\bm f_2$ are
$\bm \mu_2 =\bm K_2(\bm K_2+\sigma_2^2\bm I)^{-1}\bm y$
and $\bm \Sigma_2 = \bm K_2-\bm K_2(\bm K_2+\sigma_2^2\bm I)^{-1}\bm K_2$.

\subsection{Inference and Optimization}

In our model, we consider the hybrid prediction function $sign(a\bm f_{1}+(1-a)\bm f_{2})$, $a \in [0,1]$.
For a new point $\{\bm x^*, \bm z^*\}$, the prediction distribution $\bm f_{1*}$ of the first view and the prediction distribution $\bm f_{2*}$ of the second view are also Gaussian.
The mean of $\bm f_{1*}$ is $ {\bm k_{1*}}^ \mathbf{ T }(\bm K_1+\sigma_1^2\bm I)^{-1}\bm y$, and the covariance is
 $k_1(\bm x_*,\bm x_*)-{\bm k_{1*}}^ \mathbf{ T }(\bm K_1+\sigma_1^2\bm I)^{-1}\bm k_{1*}$.
Here, $k_1$ is the covariance function, and $\bm k_{1*}$ is the vector of covariance function values between $\bm x_*$ and the training data $\bm X$.
The mean of $\bm f_{2*}$ is
${\bm k_{2*}}^ \mathbf{ T }(\bm K_2+\sigma_2^2\bm I)^{-1}\bm y$,
and the covariance is $k_2(\bm z_*,\bm z_*)-{\bm k_{2*}}^ \mathbf{ T }(\bm K_2+\sigma_2^2\bm I)^{-1}\bm k_{2*}$.
Here, $k_2$ is the covariance function, and ${\bm k_{2*}}$ is the vector of covariance function values between $\bm z_*$ and the training data $\bm Z$.
According to the hybrid prediction function, we give our prediction on the new data point $\{\bm x^*,\bm z^*\}$.

As GPs are nonparametric models, the related hyperparameters need to be determined.
In the context of a single view, we often obtain the hyperparameters via generalized maximum likelihood. In the case of two views, we optimize the hyperparameters collaboratively by two views.
Co-regularization approaches often expect the predictions for the same observation of different views to be consistent to optimize the parameters.
Similarly, in the context of Bayesian learning, we want the posterior distributions of the latent functions of the same observation across different views to be consistent to obtain the hyperparameters, which can be realized via minimizing the above objective function shown in (\ref{eq4}).

The hyperparameters in our model can be divided into two classes: the trade-off hyperparameters $a$, $b$ and the GP related hyperparameters, which include the hyperparameters $s_f^2$, $l$ in the covariance functions and the noise hyperparameters $\sigma^2$ in the likelihood.
We use $\Theta=\{ s_{f1}^2, l_1, \sigma_1^2, s_{f2}^2, l_2, \sigma_2^2\}$ to denote hyperparameters in the second group, where $s_{f1}^2, l_1, \sigma_1^2$ are hyperparameters related to the first view, and others are hyperparameters related to the second view.
The GP related hyperparameters $\Theta$ are optimized by the gradient descent method.
As our model has an elegant formulation, the gradients with respect to $\Theta$ also have graceful forms, and the code can be easily implemented by the existing toolbox \cite{rasm}.
Following the above parameter notations, the gradient w.r.t the $s_{f1}$ is
\begin{eqnarray}
\label{eqg}
&&\frac{\partial \L_1}{\partial s_{f1}} =
\frac{a}{2} \left\{ -\bm y^ \mathbf{T}(\bm K_1+\sigma_1^2\bm I)^{-1}\frac{2\bm K_1}{s_{f1}}(\bm K_1+\sigma_1^2\bm I)^{-1}\bm y
+tr\left[(\bm K_1+\sigma_1^2\bm I)^{-1}\frac{2\bm K_1}{s_{f1}}\right] \right\} \nonumber\\
&&+ \frac{b}{2}tr\left\{
 \bm \Sigma_2^{-1} \left[\frac{2\bm K_1}{s_{f1}}-\frac{2\bm K_1}{s_{f1}}(\bm K_1+\sigma_1^2\bm I)^{-1}\bm K_1\
 -\bm K_1(\bm K_1+\sigma_1^2 \bm I)^{-1}\frac{2\bm K_1}{s_{f1}}+  \right.\right.\nonumber\\
&&\left.\bm K_1(\bm K_1+\sigma_1^2\bm I)^{-1}\frac{2\bm K_1}{s_{f1}}(\bm K_1+\sigma_1^2\bm I)^{-1}\bm K_1
\right]
- \bm \Sigma_1^{-1}\left[\frac{2\bm K_1}{s_{f1}}- \frac{2\bm K_1}{s_{f1}}(\bm K_1+\sigma_1^2\bm I)^{-1}\bm K_1\right.\nonumber\\
&&\left.\left.+\bm K_1(\bm K_1+\sigma_1^2\bm I)^{-1}\frac{2\bm K_1}{s_{f1}}(\bm K_1+\sigma_1^2\bm I)^{-1}\bm K_1
-\bm K_1(\bm K_1+\sigma_1^2 \bm I)^{-1}\frac{2\bm K_1}{s_{f1}}\right] \bm \Sigma_1^{-1} \bm \Sigma_2\right\}\nonumber\\
&&+\frac{b}{2}\left \{\left[\frac{2\bm K_1}{s_{f1}}(\bm K_1+\sigma_1^2\bm I)^{-1} \bm y
- \bm K_1(\bm K_1+\sigma_1^2\bm I)^{-1}\frac{2\bm K_1}{s_{f1}}(\bm K_1+\sigma_1^2\bm I)^{-1}\bm y\right]^ \mathbf{ T }  \right.\nonumber\\
&&(\bm \Sigma_1^{-1} +\bm \Sigma_2^{-1})(\bm \mu_1-\bm \mu_2)
- (\bm \mu_1-\bm \mu_2)^ \mathbf{ T }\bm \Sigma_1^{-1}\left[\frac{2\bm K_1}{s_{f1}}-\frac{2\bm K_1}{s_{f1}}(\bm K_1+\sigma_1^2
\bm I)^{-1}\bm K_1\right.\nonumber\\
&&\left.+ \bm K_1(\bm K_1+\sigma_1^2\bm I)^{-1}\frac{2\bm K_1}{s_{f1}}(\bm K_1+\sigma_1^2\bm I)^{-1}\bm K_1
-\bm K_1(\bm K_1+\sigma_1^2 \bm I)^{-1}\frac{2\bm K_1}{s_{f1}}\right]{\bm \Sigma_1}^{-1}\nonumber\\
&&(\bm \mu_1-\bm \mu_2)
+(\bm \mu_1-\bm \mu_2)^ \mathbf{ T }(\bm \Sigma_1^{-1}+\bm \Sigma_2^{-1})\left[\frac{2\bm K_1}{s_{f1}}(\bm K_1+\sigma_1^2\bm I)^{-1} \bm y\right.\nonumber\\
&&\left.\left.- \bm K_1(\bm K_1+\sigma_1^2\bm I)^{-1}\frac{2\bm K_1}{s_{f1}}(\bm K_1+\sigma_1^2\bm I)^{-1}\bm y\right]
\right \}
\end{eqnarray}
The gradients with respect to other hyperparameters in $\Theta$ are similar with (\ref{eqg}), and hence we omit them here.
The trade-off hyperparameters $a$ and $b$ are obtained through grid search.

We summarize MvGP1 in Algorithm \ref{algmvgp1}.

\begin{algorithm}[th]
   \caption{Multi-view GPs with Posterior Consistency}
   \label{algmvgp1}
   \begin{flushleft}
 {\bf Input:} training data $\{\bm x_i,\bm z_i,y_i\}_{i=1}^{N_1}$, test samples $\{\bm x_i^*,\bm z_i^*,y_i^*\}_{i=1}^{N_2}$. \\
 {\bf Output:} accuracy $acc$, trade-off parameters $a$, $b$, and GP related hyperparameters $\Theta$. \\
 \end{flushleft}
\begin{algorithmic}[1]
   \STATE{initialize $\Theta$ randomly.}
   \FOR{$k=1$ to $10$}
   \STATE{Divide the training data $\{\bm x_i,\bm z_i,y_i\}_{i=1}^{N_1}$ into the training set $\{\bm x_i^t,\bm z_i^t,y_i^t\}_{i=1}^{N_1^t}$ and the validation set $\{\bm x_i^v,\bm z_i^v,y_i^v\}_{i=1}^{N_1^v}$.}
   \FOR{$a$, $b$ in the search grids}
   \WHILE{termination conditions are not satisfied}
   \STATE {Update $\Theta$ by gradient descent to minimize $\L_1$ in (\ref{eq4}).}
   \ENDWHILE
   \STATE {Calculate the predictions by $sign(a\bm f_{1}+(1-a)\bm f_{2})$ on the validation set.}
   \STATE {Calculate the accuracy $acc_v$ on the validation set $\{\bm x_i^v,\bm z_i^v,y_i^v\}_{i=1}^{N_1^v}$.}
   \IF{$acc_v$ is larger than the accuracy on the last iteration}
   \STATE {record the trade-off parameters $a$, $b$. }
   \ENDIF
   \ENDFOR
   \ENDFOR
   \WHILE{termination conditions are not satisfied}
   \STATE {Update $\Theta$  by gradient descent to minimize $\L_1$ in (\ref{eq4}).}
   \ENDWHILE
   \STATE {Calculate the predictions by $sign(a\bm f_{1}+(1-a)\bm f_{2})$ on the test samples.}
   \STATE {Calculate the accuracy $acc$ on the test samples $\{\bm x_i^*,\bm z_i^*,y_i^*\}_{i=1}^{N_2}$ .}
\end{algorithmic}
\end{algorithm}



%

\subsection{Extension to Multiple Views}
In the above sections, we take two views as an example to illustrate our model MvGP1.
This model can be easily extended to multiple views because of the elegant formulation.
The posterior distribution of the latent function on each view is the Gaussian distribution. Moreover, the KL divergence \cite{joy} between two Gaussian distributions can be calculated analytically,
i.e., for two Gaussian distributions $N_0=\mathcal{N}(\bm \mu_0,\bm \Sigma_0)$ and $N_1=\mathcal{N}(\bm \mu_1, \bm \Sigma_1)$, we have
$KL(N_0||N_1)=\frac{1}{2} \{\log(\frac{|\bm\Sigma_1|}{|\bm \Sigma_0|} )+tr(\bm \Sigma_1^{-1}\bm \Sigma_0)+(\bm \mu_1-\bm\mu_0)^ \mathbf{ T }\bm \Sigma_1^{-1}(\bm \mu_1-\bm \mu_0)-N\}$. Therefore, we can extend MvGP1 to multiple views by regularizing the weighted logarithm with the KL divergences between every pair of the distinct posterior distributions.
Following the notations in MvGP1, given a data set which involves $K$ views, $p(\bm y|\bm X^k) $ denotes the marginal distribution of the $k$th view, and $p_i$ represents the corresponding posterior distribution of the latent function $\bm f_i$. The objective function in multiple views is
\begin{equation}
 \min  -\sum_{k=1}^Ka_k\log p(\bm y|\bm X^k) + \sum_{i=1}^K\sum_{j>i}^K\{b_{ij}[KL(p_i||p_j)+KL(p_j||p_i)]\}.
\end{equation}

\section{Multi-view GPs with Selective Posterior Consistency}
When the predictions from different views are not consistent on some data points,
the multi-view sufficiency assumption and compatibility assumption can not be satisfied
on the whole data set.
In this context, it is not appropriate to enforce the posteriors on the whole data set across the different views as similar as possible.
For instance, in some cases, the input data of some views may be largely affected by the noises, which may lead the predictions of these points in these views totally different from those in other views. In these cases, the multi-view assumptions fail, and enforcing the predictions of all views on these data points consistent seems to be improper.

Considering the above problem, we improve MvGP1 and present multi-view GPs with selective posterior consistency (MvGP2).
In this model, we modify the regularization term in the objective function to make the posterior distributions across different views on a subset of the data set other than the whole one as similar as possible.
In order to find the proper subset, namely the consistent set, we first optimize the hyperparameters through MvGP1 on the training set and give the predictions for the training data on each view.
Next, we select data points whose predictions on each view are all consistent and are also consistent with the true label to construct the consistent set.
Finally, we optimize the hyperparameters with the analogous procedure to MvGP1 except that we regularize the posteriors only on the chosen consistent set.

In fact, the key idea of MvGP2 is to construct the consistent set and constrain the multi-view assumptions on the consistent set. We construct the consistent set by selecting the data points whose label predictions on each view are all equal and the same as true labels.
Formally, given two views of the input data $\bm X=[\bm x_1,...,\bm x_N]^\mathbf{ T } $, $\bm Z=[\bm z_1,...,\bm z_N]^\mathbf{T} $ and the corresponding label data $\bm{y}=[y_1,...,y_N]^\mathbf{T}$, we find a index set $T=[t_1,t_2...,t_k]$, $(k<=N)$ such that for each $t\in T$, the predictions for $\bm X_t$, $\bm Z_t$ and the corresponding $\bm y_t$ are all agreed.
Then the consistent set is $\{\bm x_i,\bm z_i,y_i\}_{i\in T} $.
Let $\bm {X}_T$,$\bm {Z}_T$, and $\bm {y}_T$ denote the corresponding data matrix of the consistent set.
After constructing the consistent set, we modify the objective function to only restrict the KL divergence of the posterior distributions of the latent functions on the consistent set to be minimized,  \begin{eqnarray}
\label{eq5}
\min \L_1'= \{-[a\log p(\bm y|\bm X)+(1-a)\log p(\bm y|\bm Z)] \nonumber \\
 +\frac{b}{2}[KL({p_1}'||{p_2}')+KL({p_2}'||{p_1}')]\} ,
 \end{eqnarray}
where $\ {p_1}'= p(\bm f_1'|\bm X_T,\bm{y}_T)$
 is the posterior distribution of the latent function ${\bm f_1}'$ on the set $\{\bm {X}_T, \bm y_T\}$,
 and ${p_2}'= p({\bm f_2}'|\bm Z_T,\bm{y}_T)$
is the posterior distribution of the latent function $\bm{f_2}'$ on the set $\{\bm {Z}_T, \bm y_T\}$.

We summarize MvGP2 in Algorithm \ref{algmvgp2}.
The idea of constraining the multi-view assumption on the consistent set other than the whole data set is a novel view in multi-view learning. In the real word, data are complex and noisy. It is likely that not all the data satisfy that the predictions on different views are equal.  Moreover, we have verified this selective idea through the experiments on real word data sets in the following section.

\begin{algorithm}[th]
   \caption{Multi-view GPs with Selective Posterior Consistency}
   \label{algmvgp2}
   \begin{flushleft}
{\bf Input:} training data $\{\bm x_i,\bm z_i,y_i\}_{i=1}^{N_1}$, test samples $\{\bm x_i^*,\bm z_i^*,y_i^*\}_{i=1}^{N_2}$. \\
 {\bf Output:} accuracy $acc$, trade-off parameters $a$, $b$, GP related hyperparameters $\Theta$ and the consistent index set $T$. \\
 \end{flushleft}
\begin{algorithmic}[1]
   \STATE{ $T=\{\}$, initialize $\Theta$ randomly.}
   \STATE{run MvGP1.}
   \FOR{ $i=1$ to $N_1$}
   \STATE {Calculate the predictions by $ f_{i1}$, $ f_{i2}$ on the training data $\bm x_i$ and $\bm z_i$, respectively.}
   \IF{ $ f_{i1} =  f_{i2}$ $\&\&$ $f_{i2}=  y_i$ }
   \STATE{ add $i$ into the set $T$}
   \ENDIF
   \ENDFOR
   \FOR{$k=1$ to $10$}
   \STATE{Divide the training data $\{\bm x_i,\bm z_i,y_i\}_{i=1}^{N_1}$ into the training set $\{\bm x_i^t,\bm z_i^t,y_i^t\}_{i=1}^{N_1^t}$ and the validation set $\{\bm x_i^v,\bm z_i^v,y_i^v\}_{i=1}^{N_1^v}$.}
   \FOR{$a$, $b$ in the search grids}
   \WHILE{termination conditions are not satisfied}
   \STATE {Update $\Theta$ by gradient descent to minimize $\L_1'$ in (\ref{eq5}).}
   \ENDWHILE
   \STATE {Calculate the predictions by $sign(a\bm f_{1}+(1-a)\bm f_{2})$ on the validation set.}
   \STATE {Calculate the accuracy $acc_v$ on the validation set $\{\bm x_i^v,\bm z_i^v,y_i^v\}_{i=1}^{N_1^v}$.}
   \IF{$acc_v$ is larger than the accuracy on the last iteration}
   \STATE {record the trade-off parameters $a$, $b$. }
   \ENDIF
   \ENDFOR
   \ENDFOR
   \WHILE{termination conditions are not satisfied}
   \STATE {Update $\Theta$  by gradient descent to minimize $\L_1'$ in (\ref{eq5}).}
   \ENDWHILE
   \STATE {Calculate the predictions by $sign(a\bm f_{1}+(1-a)\bm f_{2})$ on the test samples.}
   \STATE {Calculate the accuracy $acc$ on the test samples $\{\bm x_i^*,\bm z_i^*,y_i^*\}_{i=1}^{N_2}$.}
\end{algorithmic}
\end{algorithm}

\section{Experiments}
In this section, we performed experiments with our proposed MvGP1 and MvGP2 on multiple real word data sets.
For comparison, we use three single-view methods corresponding to GPs named GP1, GP2, GP3, which use the first view, the second view, and the combination of two views, i.e., concatenating two views to construct new high-dimensional data, respectively. In addition, we also compare our algorithms with a multi-view method SVM-2K \cite{jas}, which is a two-view version of SVMs and is also inspired by the thought of co-regularization.

\subsection{Data Sets}
\subsubsection{Web-Page}
The web-page data sets have been extensively used in multi-view learning, which consist of two-view web pages collected from computer science department web sits at four universities: Cornell university, university of Washington, university of Wisconsin, and university of Texas. The two views are words occurring in a web page and words appearing in the links pointing to that page. The documents are described by 1703 words in the content view, and by 569 links between them in the cites views. We list the statistical information about the four data sets in Table~\ref{tb1}. The web pages are distributed over five classes: student, project, course, staff and faculty. We set the category with the greatest size to be the positive class (denoted as class 1), and all the other categories as the negative class (denoted as class 2) in each data set.

\begin{table}
\centering
\caption{Statistical information of four web-page data sets. }
\begin{tabular}{|c|c|c|c|c|c|c|}
\hline
Data set & size & view size & content dimension & cite dimension  & class 1 size & class 2  size\\
\hline
 Cornell & 195 & 2 & 1703 & 195 & 83 &112\\
\hline
Washington& 230 & 2 & 1703  & 230 & 107 &123\\
\hline
Wisconsin& 265 & 2 & 1703  & 265 & 122 &143 \\
\hline
  Texas& 187 & 2 & 1703 & 187 & 103 &84 \\
\hline
\end{tabular}
\label{tb1}
\end{table}

\subsubsection{Ionosphere}
Downloaded from UCI,\footnote{Data sets are available at http://archive.ics.uci.edu/ml/.} the ionosphere data set is collected by a system in Goose Bay, Labrador. This system involves a phased array of 16 high-frequency antennas with a total transmitted power on the order of 6.4 kilowatts. The targets are free electrons in the ionosphere.
Those showing evidence of some type of structure in the ionosphere are good radar returns, while those not showing the above phenomenon and whose signals pass through the ionosphere are bad returns. The data set consists of 351 examples in which 225 are "good" instances and 126 are "bad" instances.
There is only one view in this data set, and we generate the other view via principal component analysis, resulting in two views, which have 35 and 24 dimensions, respectively.

\subsection{Experimental Setting}
In the experiments, we select $60\%$ data in each data set as the training set, and the rest as the test set.
Multiple values of the hyperparameters $a$ and $b$ in MvGP1 and MvGP2 are explored in all the experiments. Given a division of the training and test set, we use cross validation with 10 folds and $20\%$ training set as the validation set for the selection of the hyperparameters $a$ and $b$ in MvGP1 and MvGP2. The considered grid ranges are  $a\in\{0, 0.1, 0.2, 0.3, 0.4, 0.5, 0.6, 0.7, 0.8, 0.9, 1 \}$ and $b\in\{ 2^{-18}, 2^{-12}, 2^{-8}, 2, 2^3, 2^8\}$. Other hyperparameters in MvGP1, MvGP2 and hyperparameters in GP1, GP2, GP3 are initialized randomly. As for the kernel function in GP1, GP2, GP3,  MvGP1, and MvGP2, we all use the squared exponential kernel as mentioned in (\ref{eq1}).
We repeat the experiments for all the data sets five times and record the average accuracies and the corresponding standard deviations.

We compared our proposed MvGP1, MvGP2 with multi-view method SVM-2K, and three single view methods GP1, GP2, and GP3.
For SVM-2K, besides the prediction functions $sign(f_1)$ and $sign(f_2)$ from the separated views, we also consider the hybrid prediction function $sign((f_1+f_2)/2)$.

\subsection{Results}
We present the average accuracies and standard deviations of all the methods on the webpage data sets and ionosphere data set in Table~\ref{tb2}.

It is clearly shown in Table~\ref{tb2} that our proposed methods MvGP1 and MvGP2 are superior to GP1, GP2, GP3 and SVM-2K. We can also observe that MvGP2 further improves the performance over MvGP1, which benefits from the idea of using the selective posterior regularization other than the posterior regularization on the whole data sets to ensure the consistency.

\begin{table}
\centering
\caption{The average accuracies and standard deviations ($\%$) of six methods on real world data sets. }
\begin{tabular}{|c|c|c|c|c|c|}
\hline
Data set & Cornell  & Washington & Wisconsin &Texas & Ionosphere  \\
\hline
GP1& 80.26$\pm$14.52 &  67.61$\pm$14.71 &72.64$\pm$16.02 & 56.01$\pm$6.32&$84.75\pm2.15$
\\
\hline
GP2&  62.56$\pm$7.87 & 74.78$\pm$1.42  & 61.51$\pm$6.27 & 73.52$\pm$12.66&$98.72\pm 1.37$
\\
\hline
GP3&  77.95$\pm$14.66 & 73.04$\pm$14.89 & 75.85$\pm$17.15 & 62.35$\pm$16.98&$97.87\pm4.76$
\\
\hline
SVM-2K& 73.68$\pm$5.04 & 74.78$\pm$4.38 &75.28$\pm$5.75& 75.15$\pm$9.44& $99.72\pm0.39$
\\
\hline
 MVGP1& 85.64 $\pm$6.87 & 86.30$\pm$6.08 & $\bm{91.32}$ $\pm$1.55 &78.33 $\pm$15.14
 & 99.29 $\pm$1.24
\\
\hline
MVGP2& $\bm{87.18}$ $\pm$5.94 & $\bm{87.18}$ $\pm$5.94 &$\bm{91.32}$ $\pm$2.04& $\bm{81.29}$ $\pm$10.81 &$ \bm{100}\pm0$
 \\
 \hline
\end{tabular}
\label{tb2}
\end{table}

%

\section{Conclusion}
In this paper, we have proposed MvGP1 which extends GPs to the scenario of learning with multiple views via the methods of posterior consistency regularization. This approach is very intuitive, resulting in an elegant objective function formulation. Experimental results on real-word web-page classification validate the effectiveness of the proposed MvGP1.
Moreover, considering that the multi-view assumptions may not be met on all data points,
we have proposed MvGP2, which constructs a consistent set and constrains the posterior consistency regularization on the consistent set other than the whole data set, leading to further improvements of the performance.
In fact, the idea of constraining the multi-view assumptions on a selective consistent set other than the whole data set is general.
It not only can be applied to GPs, but also can inspire other multi-view learning methods.

In the future, we will attempt to extend the proposed models to the scenario of big data, which may use Nystr$\rm \ddot o$m methods or other approximate approaches.

\subsubsection*{Acknowledgments.} The corresponding author Shiliang Sun would like to thank supports from the National Natural Science Foundation of China under Projects 61673179 and 61370175.


\begin{thebibliography}{10}
\providecommand{\url}[1]{\texttt{#1}}
\providecommand{\urlprefix}{URL }

\bibitem{kap}
Ashish, K., Grauman, K., Urtasun, R., Darrell, T.: {Gaussian} processes for
  object categorization. International Journal of Computer Vision  88(2),
  169--188 (2010)

\bibitem{blu}
Blum, A., Mitchell, T.: Combining labeled and unlabeled data with co-training.
  In: Proceedings of the 11th Annual Conference on Computational Learning
  Theory. pp. 92--100 (1998)

\bibitem{bon}
Bonilla, E.V., Chai, K.M.A., Williams, C.K.I.: Multi-task {Gaussian} process
  prediction. In: Proceedings of the 21th Annual Conference on Neural
  Information Processing Systems. pp. 153--160 (2007)

\bibitem{dam}
Damianou, A.C., Titsias, M.K., Lawrence, N.D.: Variational {Gaussian} process
  dynamical systems. In: Proceedings of the 25th Annual Conference on Neural
  Information Processing Systems. pp. 2510--2518 (2011)

\bibitem{eng}
Engel, Y., Mannor, S., Meir, R.: Reinforcement learning with {Gaussian}
  processes. In: Proceedings of the 22th International Conference on Machine
  Learning. pp. 201--208 (2005)

\bibitem{jas}
Farquhar, J.D.R., Hardoon, D.R., Meng, H., Shawe{-}Taylor, J., Szedm{\'{a}}k,
  S.: Two view learning: {SVM-2K}, theory and practice. In: Proceedings of the
  19th Annual Conference on Neural Information Processing Systems. pp. 355--362
  (2005)

\bibitem{joy}
Joyce, J.M.: {Kullback-Leibler} Divergence. Springer (2011)

\bibitem{kra}
Krause, A., Guestrin, C.: Nonmyopic active learning of {Gaussian} processes: an
  exploration-exploitation approach. In: Proceedings of the 24th International
  Conference on Machine Learning. pp. 449--456 (2007)

\bibitem{law}
Lawrence, N.D., Jordan, M.I.: Semi-supervised learning via {Gaussian}
  processes. In: Proceedings of the 18th Annual Conference on Neural
  Information Processing Systems. pp. 753--760 (2004)

\bibitem{ras}
Rasmussen, C.E., Williams, C.K.I.: {Gaussian} Processes for Machine Learning.
  MIT Press (2006)

\bibitem{rasa}
Rasmussen, C.E., Kuss, M.: {Gaussian} processes in reinforcement learning. In:
  Proceedings of the 17th Annual Conference on Neural Information Processing
  Systems. pp. 751--758 (2003)

\bibitem{rasm}
Rasmussen, C.E., Nickisch, H.: {Gaussian processes for machine learning (GPML)
  toolbox}. Journal of Machine Learning Research  11(Nov),  3011--3015 (2010)

\bibitem{sho}
Shon, A.P., Grochow, K., Hertzmann, A., Rao, R.P.N.: Learning shared latent
  structure for image synthesis and robotic imitation. In: Proceedings of the
  19th Annual Conference on Neural Information Processing Systems. pp.
  1233--1240 (2005)

\bibitem{sin}
Sindhwani, V., Chu, W., Keerthi, S.S.: Semi-supervised {Gaussian} process
  classifiers. In: Proceedings of the 20th International Joint Conference on
  Artificial Intelligence. pp. 1059--1064 (2007)

\bibitem{sun}
Sun, S.: A survey of multi-view machine learning. Neural Computing and
  Applications  23(7-8),  2031--2038 (2013)

\bibitem{sun2}
Sun, S., Jin, F.: Robust co-training. International Journal of Pattern
  Recognition and Artificial Intelligence  25(07),  1113--1126 (2011)

\bibitem{xuc}
Xu, C., Tao, D., Li, Y., Xu, C.: Large-margin multi-view {Gaussian} process for
  image classification. In: Proceedings of the 5th International Conference on
  Internet Multimedia Computing and Service. pp. 7--12. ACM (2013)

\bibitem{xu}
Xu, C., Tao, D., Xu, C.: A survey on multi-view learning. arXiv preprint
  arXiv:1304.5634  (2013)

\bibitem{yu}
Yu, K., Tresp, V., Schwaighofer, A.: Learning {Gaussian} processes from
  multiple tasks. In: Proceedings of the 22th International Conference on
  Machine Learning. pp. 1012--1019 (2005)

\bibitem{yus}
Yu, S., Krishnapuram, B., Rosales, R., Rao, R.B.: Bayesian co-training. Journal
  of Machine Learning Research  12,  2649--2680 (2011)

\bibitem{zha}
Zhao, J., Sun, S.: Variational dependent multi-output {Gaussian} process
  dynamical systems. Journal of Machine Learning Research  17,  1--36 (2016)

\bibitem{zhou}
Zhou, J., Sun, S.: {Gaussian} process versus margin sampling active learning.
  Neurocomputing  167,  122--131 (2015)

\bibitem{zhu}
Zhu, X., Ghahramani, Z., Lafferty, J.D.: Semi-supervised learning using
  {Gaussian} fields and harmonic functions. In: Proceedings of the 20th
  International Conference on Machine Learning. pp. 912--919 (2003)

\end{thebibliography}

\end{document}